\documentclass[fleqn,twoside]{article}
\usepackage[headings]{espcrc2}

\readRCS
$Id: espcrc2.tex,v 1.2 2004/02/24 11:22:11 spepping Exp $
\usepackage{graphicx, balance}
\usepackage[figuresright]{rotating}

\usepackage{amsmath}
\usepackage{array}
\usepackage{algorithm}
\usepackage{algpseudocode}
\usepackage{subfig}
\usepackage{multirow}
\usepackage[justification=centering]{caption}

\newcommand{\xdownarrow}[1]{%
  {\left\downarrow\vbox to #1{}\right.\kern-\nulldelimiterspace}
}
\newcommand{\xuparrow}[1]{%
  {\left\uparrow\vbox to #1{}\right.\kern-\nulldelimiterspace}
}

\title{\textbf{New Fuzzy LBP Features for Face Recognition}}

\author{Abdullah Gubbi\address[DCSE]{Department of Electronics and Communications, P.A. College of Engineering, Mangalore, India, Contact: abdullahgubbi@yahoo.com \\},
Mohammed Fazle Azeem\address{Department of Electrical Engineering, Aligarh Muslim University, Aligarh, India, \qquad\qquad Contact: mf.azeem@gmail.com \\}
Zahid Ansari\address{Department of Computer Science, P.A. College of Engineering, Mangalore, India, \qquad\qquad Contact: zahid\_cs@pace.edu.in}}

\runtitle{New Fuzzy LBP Features for Face Recognition}
\runauthor{Abdullah Gubbi, et al.,}

\begin{document}
\begin{abstract}

There are many Local texture features each very in way they implement and each of the Algorithm trying improve the performance. An attempt is made in this paper to represent a theoretically very simple and computationally effective approach for face recognition. In our implementation the face image is divided into 3x3 sub-regions from which the features are extracted using the Local Binary Pattern (LBP) over a window, fuzzy membership function and at the central pixel. The LBP features possess the texture discriminative property and their computational cost is very low. By utilising the information from LBP, membership function, and central pixel, the limitations of traditional LBP is eliminated. The bench mark database like ORL and Sheffield Databases are used for the evaluation of proposed features with SVM classifier.  For the proposed approach K-fold and ROC curves are obtained and results are compared.  
\end{abstract}

\maketitle

\section{INTRODUCTION}
Face Recognition (FR) is an effortless task for humans while it is difficult task for machines due to pose and illumination variation, ageing, facial growth to mention a few. While having so many difficulties face carries more information as compared to iris, figure, gait etc.  FR has advantage that, it can be captured at a distance and in hidden manner whereas other biometric modalities like fingerprint, palm prints and knuckle prints require physical contact with a sensor. The FR field is active research area because of surveillance, Pass port, ATM etc. The problem with FR is the high-dimensionality of feature space (original space). A straightforward implementation with original image space is computationally costly. Therefore, techniques of feature extraction and feature selection are used. In the recent past many FR algorithms have been implemented. These algorithms can be roughly classified into three categories:
\begin{enumerate}
\item Feature based matching methods that deals with local features such as the eyes, mouth, and nose and their statistics which are fed into the recognition system [14].
\item Appearance-based schemes that make use of the holistic texture features [17].
\item Hybrid Features that uses local and holistic features for the recognition. It is reported that the Hybrid approach gives better recognition results [26].
\end{enumerate} 
One of the implementations of the structure-based schemes is based on the human perception [24].
 $ \; $In this implementation, a set of geometric face features, such as eyes, nose, mouth corners etc., is extracted. The features are selected based on the specific facial points chosen by a human expert. The positions of the different facial features form a feature vector which is the input to a structural classifier to identify the subject.
In the Appearance-based schemes, two of the most popular techniques are Principal Component Analysis (PCA) [20]
 and Linear Discriminant Analysis (LDA) [27].

\textit{PCA}: Principle Component Analysis is one of the holistic approaches. This is one of the statistical approaches based on unsupervised learning method. In which the Eigen vectors are computed from the covariance matrix.  PCA  model the linear variations in high  dimensional data. PCA preserves the global structure of the image in the sense the image can be completely reconstructed with PCA. It does the guarantees dimensionality reduction by projecting the $n$-dimensional data onto the $r$ ($r<<n$) dimensional linear subspace. The objective here is to find a set of mutually orthogonal basis functions that capture the directions of maximum variance in the data. Since PCA  is used for  the dimensionality reduction, hence it can be used for image compression. We can find many holistic approaches in the literature like kernel PCA, 2D-PCA Complex PCA etc

\textit{LDA}: Linear Discriminant Analysis, L is called supervised learning algorithm method. The Fisher faces method he build in the concept of LDA. LDA builds the projection axis in such a way that the data points belonging to same class are nearer. and  projection points are for  the interclass. LDA encodes the discriminating information in a linear separable space using bases which are not necessarily orthogonal unlike in the case of PCA. As the face samples involve high dimensional image space, the Eigen-faces approach is likely to find the wrong components that leads to poor recognition. on the other hand, the Fisher-faces are computationally expensive.

The ultimate aim of any feature extraction algorithm is, it should build highly describable features. The developed features should produce ideally zero variance within the class. and should produce infinite variance for the interclass.

The organization of the rest of paper is as follows. Section 2 describes the basic LBP and it's variants. Section 3 discusses the fuzzy logic. In section 4 the proposed method is presented. Section 5 describes the experimental setup and results. Section 6 gives the conclusions of this paper.

\section{BASIC LBP AND ITS VARIANTS}

In this section brief concepts of basic LBP and it’s Variants are discussed The basic LBP was proposed by [15]
 and some of it’s variants are Tan and Triggs [19]
  $ \; $proposed Local Ternary Pattern (LTP) uses three levels (+1, 0,-1), these three level are obtained by quantizing the difference between a  central pixel and its neighbouring pixel gray. Some of the variants of LBP, are derivative-based LBP [9]
  , Sobel-LBP [25]
  , Uniform LBP [2]
  , Rotation invariant LBP [5]
  , Multi-dimensional LBP [18]
  , dominant LBP [13]
  , centre-symmetric LBP [21]
   $ \; $and Transition LBP [1]
   , have been proposed .
\subsection{Basic Local Binary Patterns}  
\begin{figure}[t]
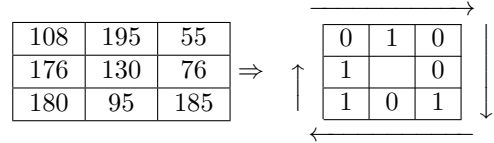

\begin{centering}
$
 \begin{tabular}{|c|c|c|}
 \hline
 108 &	195	& 55 \\
 \hline
 176 &	130	& 76 \\
 \hline
 180 &	95	& 185 \\
 \hline
 \end{tabular}
\Rightarrow
$
\begin{tabular}{c|c|c|c|c}
\multicolumn{5}{c}{$\xrightarrow{\hspace*{2cm}}$} \\
\cline{2-4}
& 0 & 1 & 0 & \multirow{3}{*}{$\xdownarrow{7mm}$}\\
\cline{2-4}
\multirow{2}{*}{$\xuparrow{4mm}$} & 1 &  & 0 & \\
\cline{2-4}
& 1 & 0 & 1 & \\
\cline{2-4}
\multicolumn{5}{c}{$\xleftarrow{\hspace*{2cm}}$} \\
\end{tabular}
\caption{The basic Local Binary Pattern (LBP) operator}
\label{fig:BasicLBPOperator}
\end{centering}
\end{figure}

LBP concept is applied to area like face recognition [1]
, dynamic texture recognition [23]
 $ \; $and shape localization [9].
 $ \; $ The Local Binary Pattern (LBP) method is widely used in 2D texture analysis. The LBP operator is a non-parametric 3x3 kernel which describes the local spatial structure of an image. It was first introduced by Ojala et al [15]
 $ \; $ who showed the high discriminative power of this operator for texture classification. At a given pixel position $(x_c,y_c )$, LBP is defined as an ordered set of binary comparisons of pixel intensities between the centre pixel and its eight surrounding pixels. The decimal values of the resulting 8-bit word (LBP code) leads to 28 possible combinations, which are called Local Binary Patterns abbreviated as LBP codes with the 8 surrounding pixels. The basic LBP operator is a fixed $3 \times 3$ neighbourhood as in Fig. \ref{fig:BasicLBPOperator}. 

The LBP operator can be mathematically expressed as:
\begin{equation}
LBP(x_c, y_c)=\sum_{n=1}^8 s(I_n - I_c) 2^n
\end{equation}
Where $I_c$ corresponds to the gray value of the centre pixel $(x_c, y_c)$, $I_n$ to the gray values of the 8 surrounding pixels and function  s is defined as,
\begin{equation}\label{eq2}
s(x) \leftarrow
\begin{cases}
1, & \text{if } \: x \geq 0 \\
0, & \text{if } \: x < 0
\end{cases}
\end{equation}

\subsection{Uniform LBP} 
Uniform LBP binary patterns, proposed by T. Ojala et. al [16]
$ \, $  have certain fundamental properties of texture. In this implementation they have considered if the frequency of occurrence exceeds 90\% then they call it as uniform patterns. The frequency considered here is the transition from zero to one and one to zero.  
\subsection{Multi-block LBP (MB-LBP)}  
 As in the case of basic LBP the central pixel is taken as threshold and neighbouring pixel is assigned values. In the case of MB-LBP [22]
 $ \; $the same procedure is adapted. They consider the rectangular box as the reference. For this reference rectangular box the average value is computed and this average value will be acting as threshed, and the neighbouring blocks are assigned the value based on threshold.
\subsection{3-Patch (3P-LBP) and 4-Patch (4P-LBP) LBP codes}  
3P-LBP [3]
$ \, $ is obtained by comparing the values of three neighbouring windows to produce a one bit value. A $3 \times 3$ window centred on a pixel and $S$ additional windows distributed uniformly in a ring of radius r around each pixel is considered. Pairs of windows are compared with the centre patch, and a bit set describing which of the two windows is more similar to the centre patch. In 4P-LBP, two rings of radii $r_1$ and $r_2$ are considered. The code is then produced by comparing the two centre symmetric windows in the inner ring with two centre symmetric windows in the outer ring positioned $\alpha$ patches away along the circle.
\subsection{LBP variance (LBPV)}   
In LBP variance (LBPV) [5]
, the contrast of the image is used along with neighbouring points and radii, are used to generate joint distribution of local pattern that yields a dominant texture descriptor. LBPV contains both local pattern and local contrast information. An alternative is the use of a hybrid scheme, LBP variance (LBPV), which uses globally rotation invariant matching with locally variant LBP texture features. LBPV provides efficient joint LBP and contrast distribution where the variance is used as an adaptive weight to adjust the contribution of the LBP code in histogram calculation.
In spite of the the great success of LBP in pattern recognition, its underlying working mechanism still needs more investigation. Even after having so many versions of LBP the problem still remain to be address other possibilities which can cater better recognition rate in other words which information is missing in the LBP and how to effectively represent the missing information in the LBP style so that better texture classification can be achieved?

Looking at the Limitation of the PCA, LDA  techniques and taking advantage of Information sets, which have been developed by Hanmandlu [6]
$ \, $ to enlarge the scope of fuzzy sets. Information set concept is simple understand. The image after portioned is treated as the information source. The information sources contained in windows form the fuzzy sets. The distribution of these information sources in the fuzzy sets requires an appropriate membership function to fit. When fuzzy set coupled with membership function forms the information set. We concentrate on extracting the local fuzzy features from images, instead of considering complete image which is a high-dimensional vector. To extract the simple and robust local features of an image, we use basic Local Binary Pattern (LBP), information set and the central pixel. These features are tested empirically on SVM [4].

\section{FUZZY LOGIC}

Fuzzy Logic theory has been extensively used in all the fields.  it is the extension of crisp set theory and fuzzy logic (FL) will helps to deal with imprecise and uncertain data. It tackles the concept of partial truth. It was introduced by Prof. Zadeh to model the vagueness and ambiguity in complex systems for which there is no mathematical model. 
The limitation of fuzzy sets is that information source values and their membership function values are treated separately for all the problems dealing with the fuzzy logic theory. The membership function value gives the degree of association of a particular information source value in a fuzzy set [6].
$ \; $ The proposed work seeks to combine the information source values with their membership values together called Information set. The fuzzifier developed by Hanmandlu et.al [8]
 $ \, $has the scope to enlarge the fuzzy set. To give an example of agent, consider a set of students graded based on the performance of the class topper as the benchmark and the student’s individual performance (information source value) is determined by comparing his performance with that of the topper ($\mu$).
Some of the popular membership functions are as given below.

\begin{figure}[t]
\begin{center}
    \subfloat[S Shaped Membership Function\label{subfig-1:SMF}]{%
      \centerline{\includegraphics[scale=0.35]{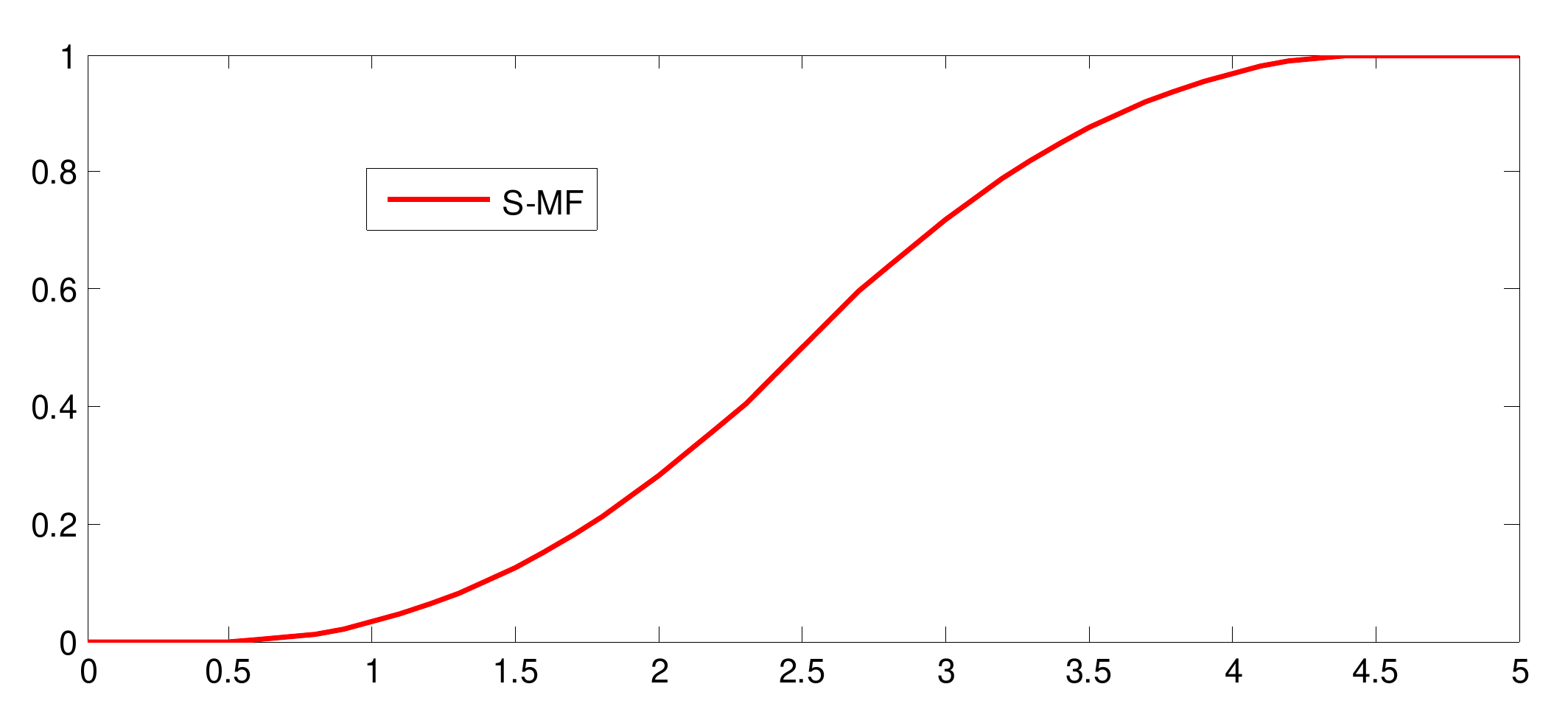}}
    }
    \newline
    \subfloat[Z Shaped Membership Function\label{subfig-2:ZMF}]{%
      \centerline{\includegraphics[scale=0.35]{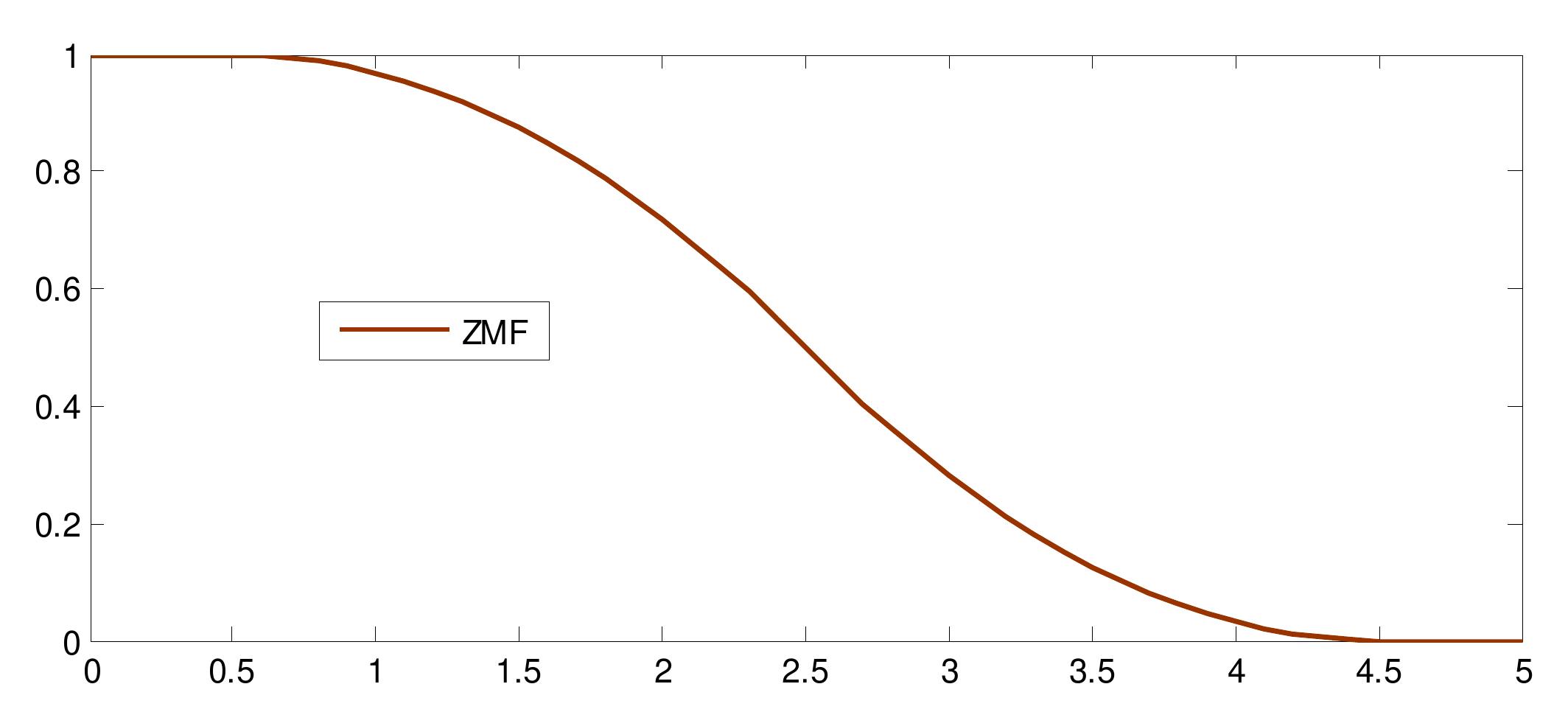}}
    }
    \newline
    \subfloat[Gaussian Membership Function\label{subfig-3:Gaussian}]{%
      \centerline{\includegraphics[scale=0.35]{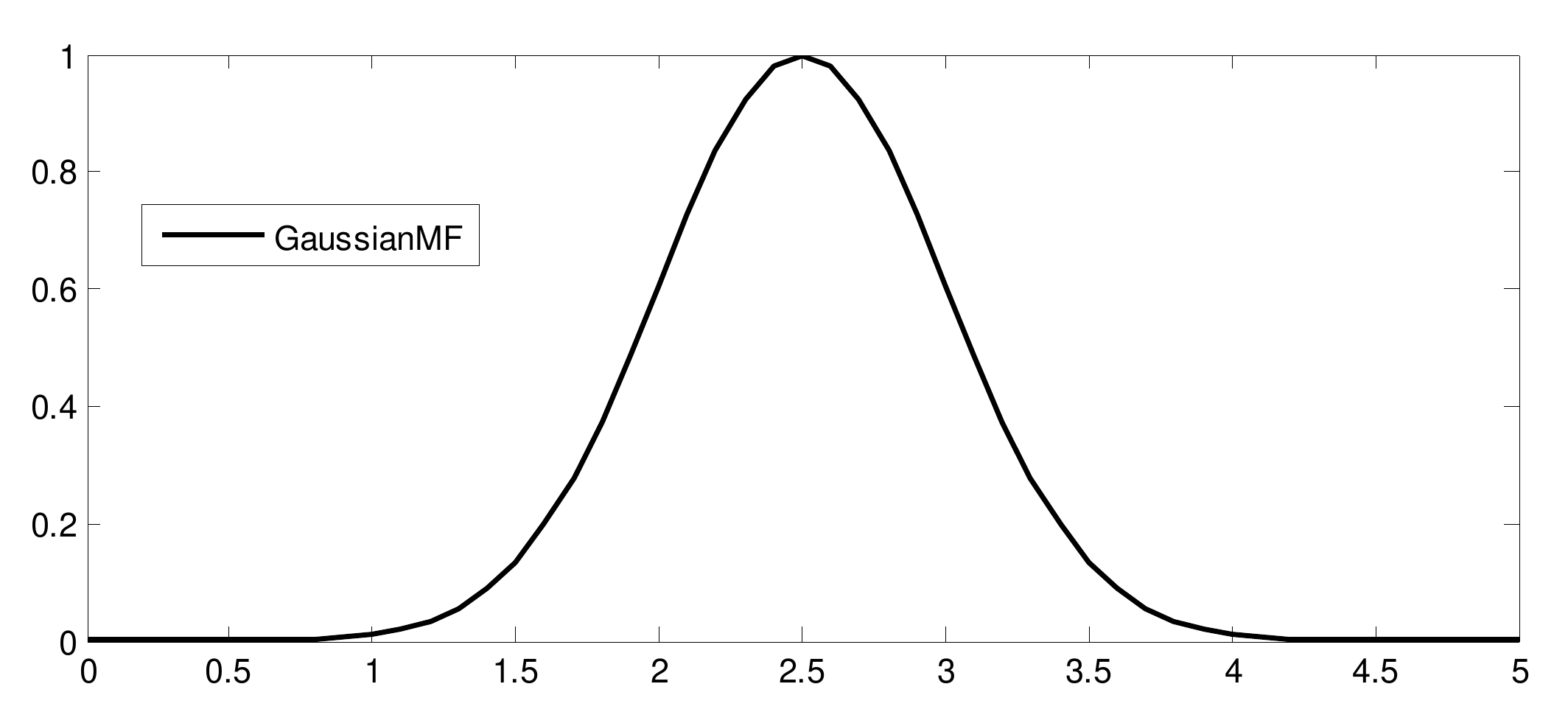}}
    }
    \caption{Fuzzy Membership Functions}
    \label{fig:MF}
\end{center}
\end{figure}

\begin{enumerate}
\item S Shaped Membership Function
is a spline-based curve. The parameters a and b locate the extremes of the sloped portion of the curve as given in equation \eqref{eq:SMF}. Fig. \ref{subfig-1:SMF} illustrates S shaped membership function.

\begin{equation}\label{eq:SMF}
\mu(x; a, b) \leftarrow
\begin{cases}
0, & \text{if } \: x \leq a \\
2\left( \frac{x-a}{b-a}\right)^2, & \text{if } \: a \leq x \leq \frac{a+b}{2} \\
1-2\left( \frac{x-b}{b-a}\right)^2, & \text{if } \: \frac{a+b}{2} \leq x \leq b \\
1, & \text{if } \: x \geq b
\end{cases}
\end{equation}


\item Z Shaped Membership Function is a spline-based function of x. The parameters a and b locate the extremes of the sloped portion of the curve as given in equation \eqref{eq:ZMF}. Fig. \ref{subfig-2:ZMF} illustrates Z shaped membership function.

\begin{equation}\label{eq:ZMF}
\mu(x; a, b) \leftarrow
\begin{cases}
1, & \text{if } \: x \leq a \\
1-2\left( \frac{x-a}{b-a}\right)^2, & \text{if } \: a \leq x \leq \frac{a+b}{2} \\
2\left( \frac{x-b}{b-a}\right)^2, & \text{if } \: \frac{a+b}{2} \leq x \leq b \\
0, & \text{if } \: x \geq b
\end{cases}
\end{equation}


\item Gaussian Membership Function
The symmetric Gaussian function depends on two parameters $\sigma$ and $c$ as given in equation \eqref{eq:Gaussian}.
\begin{equation}\label{eq:Gaussian}
\mu(x; \sigma, c) \leftarrow e^{\frac{-(x - c)^2}{2\sigma^2}}
\end{equation}
where $\sigma$ is variance and $c$ is center. These two parameter define the shape of the Gaussian bell. Fig. \ref{subfig-3:Gaussian} illustrates Gaussian membership function.


\end{enumerate}

\section{PROPOSED METHOD}

Uncertainty appears in the gray values of images. Instead of considering the whole image and their gray values, we divide the image into sub images of size $3 \times 3$ (enclosed in the window) and extract local information using LBP method. We consider non overlapping sub images. We compute a membership function for each window using S-Membership Function, Z-Membership Fuction, Gaussian Membership Function, the New proposed membership Function and Root Mean features. For all the features we use central pixel to compute the final feature, the reason being the central pixel value is lost once we obtain the LBP feature. 
Let $\mu_{i,j}$ be the membership value of the window using any of the above method the size of the membership will be $3 \times 3$ when we get this from the $3 \times 3$ window. The procedure is explained in the later part of this section. Let S be the $3 \times 3$ window in the image domain. Let $H_w$ be the information set obtained from the window, that is the element by element product of the $S$ and $\mu_{i,j}$ 	denoted as 
$H_w=S\,\text{.X}\,\mu_{i,j}$, where .X represents the element by element multiplication. After element by element multiplication, we get $3 \times 3$ matrix. we take the sum as given below.
\begin{equation}\label{eq:Hws}
H_w^s = \sum_{i=1}^w H_w 
\end{equation}
where $H_w^s$ is the information set obtained after the summation.
Once we obtain the $H_w^s$, next we compute the LBP code the example is as given in Fig. \ref{fig:BasicLBPOperator}.
In order to take account of the information from the neighbourhood pixels (information sources), we compute the LBP value for the window. This is given by:
\begin{equation}\label{eq:LW}
L_w=LBP(x_c, y_c)
\end{equation}
We convert the LBP code to the decimal value. Let the value of LBP from a window be denoted by $L_w$. The New membership function based feature is given by
\begin{equation}\label{eq:Fwk}
F_w^k = H_w^s \times L_w \times I_c     
\end{equation}
where $I_c$ represent the centre pixel in the window.
The New membership function is computed from the information of local structural details like average of the window and maximum in the window under consideration. The New membership function is given by 
\begin{equation}\label{eq:NewMu}
\mu_{i,j}=\frac{\vert I_{i,j}-I_{avg} \vert}{I_{max}}
\end{equation}
where $I_{i,j}$ represent the normalized intensity value which we will obtain by dividing the maximum in the image. $I_{avg}$ is the average in the window under consideration and $I_{max}$ is the maximum in the window under consideration.
Algorithm \ref{NEWMF} describes various steps involved for the computation of features using various membership functions for a given image.
\begin{algorithm}
{\footnotesize
\caption{Feature Extraction}\label{NEWMF}
\begin{algorithmic}[0]
\State  \textbf{Input:} Face Image $I$
\State  \textbf{Output:} New membership function based feature vector $\textbf{F}^{I}$ for image $I$
\end{algorithmic}
\begin{algorithmic}[1]
\State Normalize the image $I$ by dividing each pixel value with the maximum intensity value
\State Partition the image $I$ into $3 \times 3$ windows $I = \lbrace w^1, w^2, \cdots w^d\rbrace $
\For{$k \leftarrow 1, d$}
	\Statex // Compute the proposed new membership function values for  $w^k$
	\State $I_c = w^k_{22}$ // Central pixel value
	\For{$i \leftarrow 1,3$}
		\For{$j \leftarrow 1,3$}
			\State Compute the membership value $\mu_{ij}$ using \eqref{eq:SMF},
			 \eqref{eq:ZMF}, \eqref{eq:Gaussian} or \eqref{eq:NewMu}.
		\EndFor	
	\EndFor
	\State Compute Information Set value $H_w^s$ using \eqref{eq:Hws}
	\State Compute LBP value $L_{w^k}$ using \eqref{eq:LW}
	\State Computer the feature $F^k_w$ using \eqref{eq:Fwk}
	\State $\textbf{F}^I_k = F^k_w$ // Store in feature vector $\textbf{F}^I$
\EndFor
\end{algorithmic}
}
\end{algorithm}

\subsection{The Root Mean Square Feature} 
This feature depend upon the two parameters on is the fuzzifier and the central pixel value and the LBP code. The fuzzifier developed by Prof. Hanmandlu has the scope to expand the fuzzy set. The fuzzifier $f_h^2$ in \eqref{eq:fuzzifier} is devised by Hanmandlu et al. in [7]
 $ \, $and it gives the spread of attribute values with respect to the chosen reference (symbolized as ref). It is defined as
\begin{equation}\label{eq:fuzzifier}
f_h^2=\frac{\sum_{i=1}^w \sum_{j=1}^w(I_{ref}-I_{ij} )^4}{\sum_{i=1}^w \sum_{j=1}^w(I_{ref}-I_{ij} )^2}
\end{equation}
$I_{ref}$ can be taken as average or minimum or maximum from the window under consideration. It may be noted that the above fuzzifier gives more spread than is possible with variance as used in the Gaussian function.
\begin{equation}
F_w^k=\left(\sqrt{I_c^2+f_h^2}\right) \times I_c \times L_w
\end{equation}  

The contribution of this method is that it eliminates the shortcoming of LBP approach which ignores the central pixel value and it considers information of both central and neighbourhood pixels. 

\begin{figure*}[t]
\begin{center}
    \subfloat[ORL database images of different subjects\label{subfig-2:ORLImages}]{%
      \centerline{\includegraphics[scale=0.55]{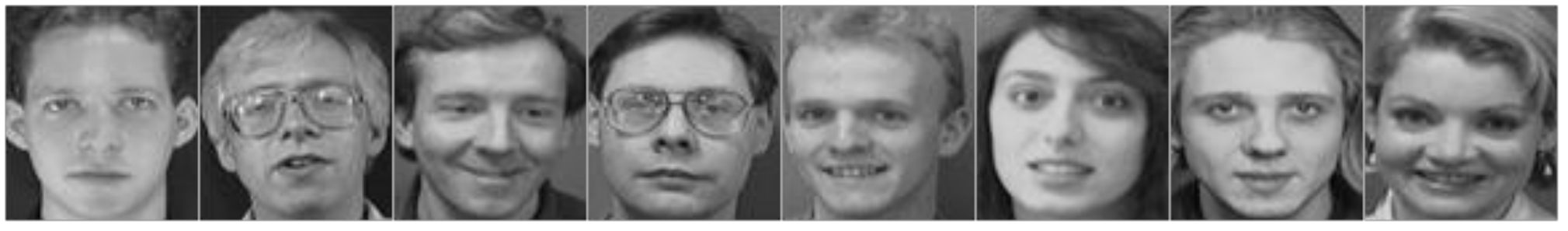}}
    }
    \newline
    \subfloat[Sheffield database images of different subject\label{subfig-1:SheffieldImages}]{%
      \centerline{\includegraphics[scale=0.46]{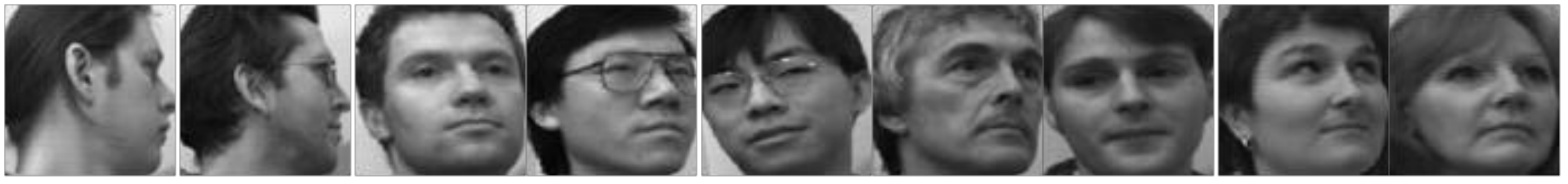}}
    }
    \caption{Sample images of benchmark databases}
    \label{fig:database}
\end{center}
\end{figure*}

\section{EXPERIMENTAL RESULTS}
The proposed method is simple and computationally efficient. A large number of experiments were conducted on ORL and Sheffield databases, shows the effectiveness of our method. Olivetti Research Laboratory (ORL) in Cambridge, U.K. AT\&T database has 400 images with 40 different subjects with 10 images per subject of size $112\times 92$. Sheffield database has 20 classes the size of the each image is approximately $220\times 220$. With lot of orientation of head, the degree of rotation is $\pm 90$ about the center axis. The 20 subjects of Sheffield database are kept in separate folders. Number of images in each older varies in number. In both databases, we normalize all of the gray (pixel) values by dividing them by the maximum gray value in that image. So the range of the pixel values lies in the range 0 to 1 (also known as $I_{i,j}$, the normalized information source values). For both databases we have resized the images to integer multiplication of 3 so as to avoid the padding of zeros at the end.

\begin{table*}[!t]
\begin{center}
\caption{Recognition rate (in \%) for ORL database with SVM classifiers}\label{tab:ORL}
\begin{tabular}{|l|c|c|c|c|c|}
\hline 
\multirow{2}{*}{ORL Database} & \multirow{2}{*}{SVM Poly1}  & \multirow{2}{*}{SVM Poly2}  & \multicolumn{3}{ |c| }{K-Fold Results} \\ \cline{4-6}
& & & {MIN} & {MAX} & {AVG} \\
\hline
\hline
SMF	& 84	& 78.5	& 87.5	& 100	& 95.25 \\
\hline
ZMF	& 82	& 79	& 87	& 100	& 94.50 \\
\hline 
Gauss MF	& 85	& 80	& 87.5	& 100	& 75.75 \\
\hline 
New MF	& 85.5	& 81	& 87.5	& 100	& 95.75 \\
\hline 
RMS	& 85	& 80.5	& 90	& 100	& 96 \\
\hline 
\end{tabular}
\end{center}
\end{table*}

\begin{figure*}[t]
\begin{centering}
\centerline{\includegraphics[scale=0.3]{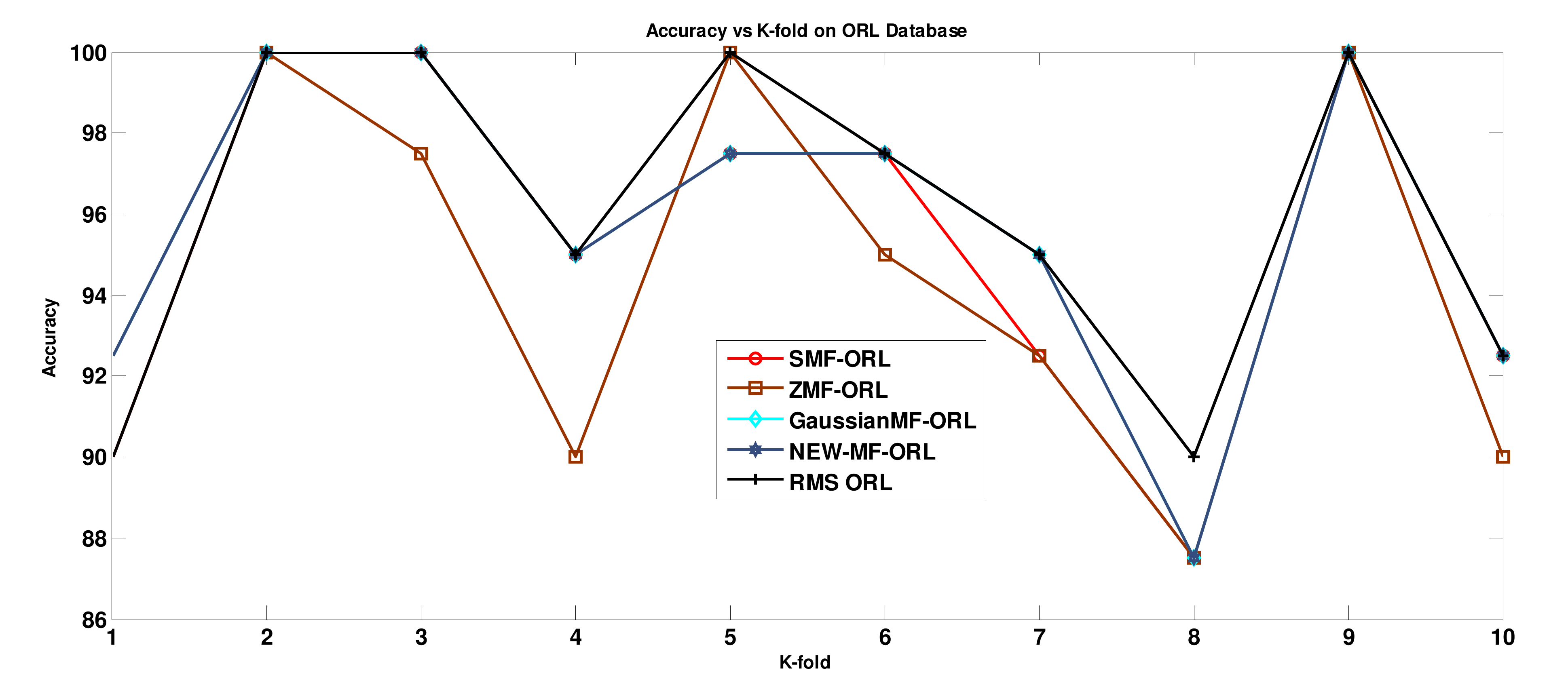}}
\caption{K-fold Recognition rates on ORL Database with 10 folds}
\label{fig:KFOLD_ORL}
\end{centering}
\end{figure*}

\begin{figure*}[t]
\begin{centering}
\centerline{\includegraphics[scale=0.3]{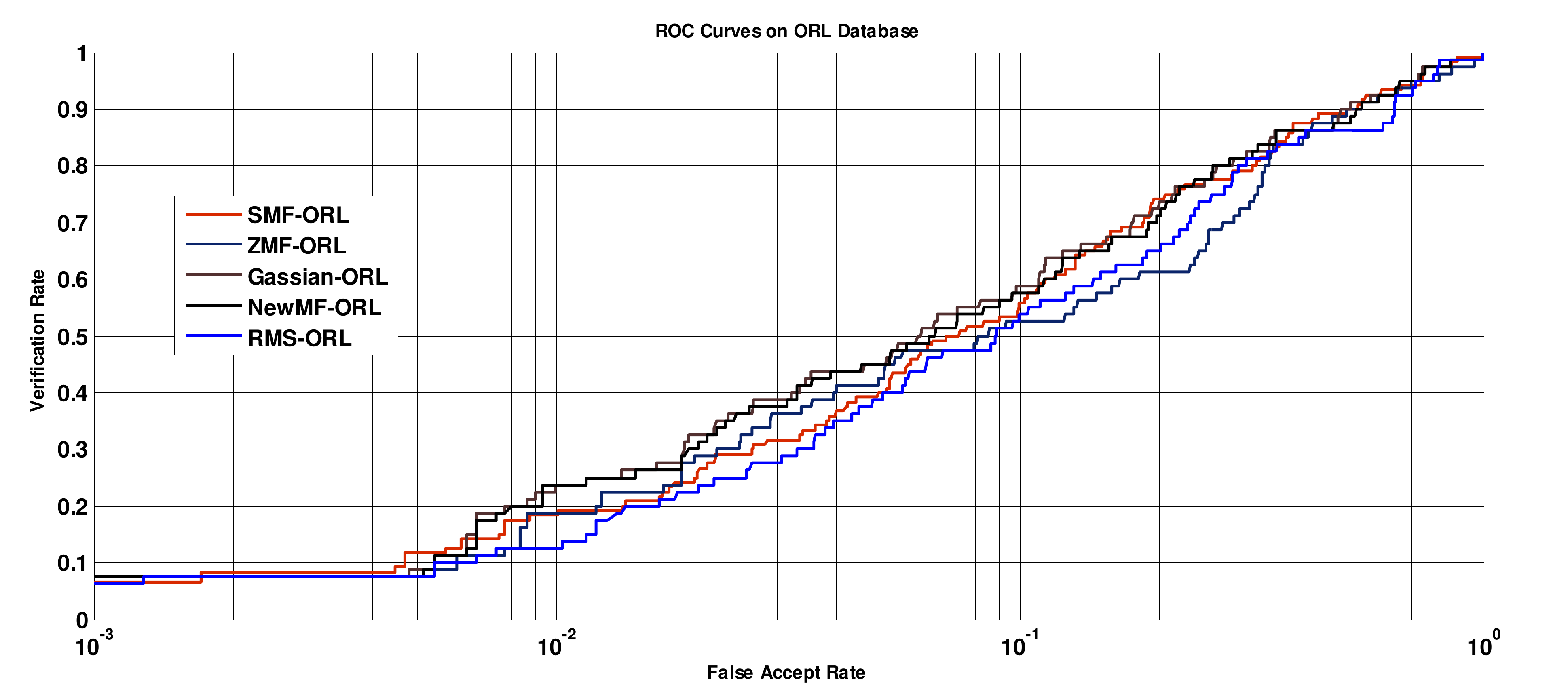}}
\caption{ROC Curves for various features on ORL database}
\label{fig:ROC_ORL}
\end{centering}
\end{figure*}

Then the next step is to divide the image into $3 \times 3$ windows which are non-overlapping. For each $3 \times 3$ window, we compute the $\mu_{i,j}$  and perform the element by element  multiplication with $S$. and obtain the  centre pixel value $i_c$  of a window, We then compute the LBP value from a window using \eqref{eq:LW}. The total information (feature) is the product of information set, the central pixel, and the decimal value obtained from the  neighbouring pixels. This total information forms a feature from each window.  All features are stored as an array for the complete image. Suppose if image is of size $63 \times 63$ we get an array of 441 features. For 400 images we store features as $400 \times 441$. The complete set is divided into the training set and the test set. We have used SVM [4]
 classifier which is available from PRT tools. The database is divided into train set and the test set. The first 50\% dataset is used for training and remaining  50\% is used for testing. For ORL database the recognition rates are given in Table \ref{tab:ORL}. The good performance is achieved from New Membership Function for all the degree of Polynomial 1,2,3. Fuzzy logic gives the flexibility of selecting membership. Either it can be the Standard Membership functions like Gaussian or sigmoid or one can design his/her own membership function to cater the need. k-fold is one of validation technique in which the data is approximately equally divided. Each time the one set of data is used for testing and the remaining portion of the data is used for training, the data which was in test will be placed back as training. Normally the procedure is repeated K times hence the name K-fold. The value of K is normally taken as 10. For our implementation we have taken SVM classifier for K-fold Validation. Fig. \ref{fig:KFOLD_ORL} shows the K-fold results of the recognition on ORL Database. The best performance is given by Root Mean Square (RMS) average recognition rate of 96\% followed by Gaussian Membership function, and the proposed New MF with average recognition rate of 95.75\%. Receiver operating characteristic (ROC) curve is graphical plot that defines the performance of a classifier system as its discrimination threshold is varied. For obtaining the ROC curves we have taken the K-NN Classifier. Fig. \ref{fig:ROC_ORL} shows the ROC results of the recognition on ORL Database. The recognition rates at 0.1 of FAR are as follows with SMF the recognition rate we got is 55\%,with ZMF the recognition rate 52\%, with Gaussian-MF the recognition rate 59\%, with New-MF the recognition rate is 57\%, and with RMS features the recognition rate obtained is 54\%. And The best performance is given by Gaussian MF and the least performance is given by ZMF.


%

\begin{table*}[!t]
\begin{center}
\caption{Recognition rate (in \%) for Sheffield database with SVM classifier}\label{tab:Sheffield}
\begin{tabular}{|l|c|c|c|c|c|}
\hline 
\multirow{2}{*}{Sheffield Database} & \multirow{2}{*}{SVM Poly1}  & \multirow{2}{*}{SVM Poly2}  & \multicolumn{3}{ |c| }{K-Fold Results} \\ \cline{4-6}
& & & {MIN} & {MAX} & {AVG} \\
\hline
\hline
SMF &	92.14 &	85.71 &	90 &	100 &	99 \\
\hline 
ZMF	& 92.14 &	87.85 &	100 &	100 &	100 \\
\hline 
Gauss MF &	95 &	92.14  &	100 &	100 &	100 \\
\hline 
New MF	& 95 &	92.14  &	100 &	100 &	100 \\
\hline 
RMS	& 97.14 &	90.71  &	90 &	100 &	99 \\
\hline 
\end{tabular}
\end{center}
\end{table*}

\begin{figure*}[t]
\begin{centering}
\centerline{\includegraphics[scale=0.3]{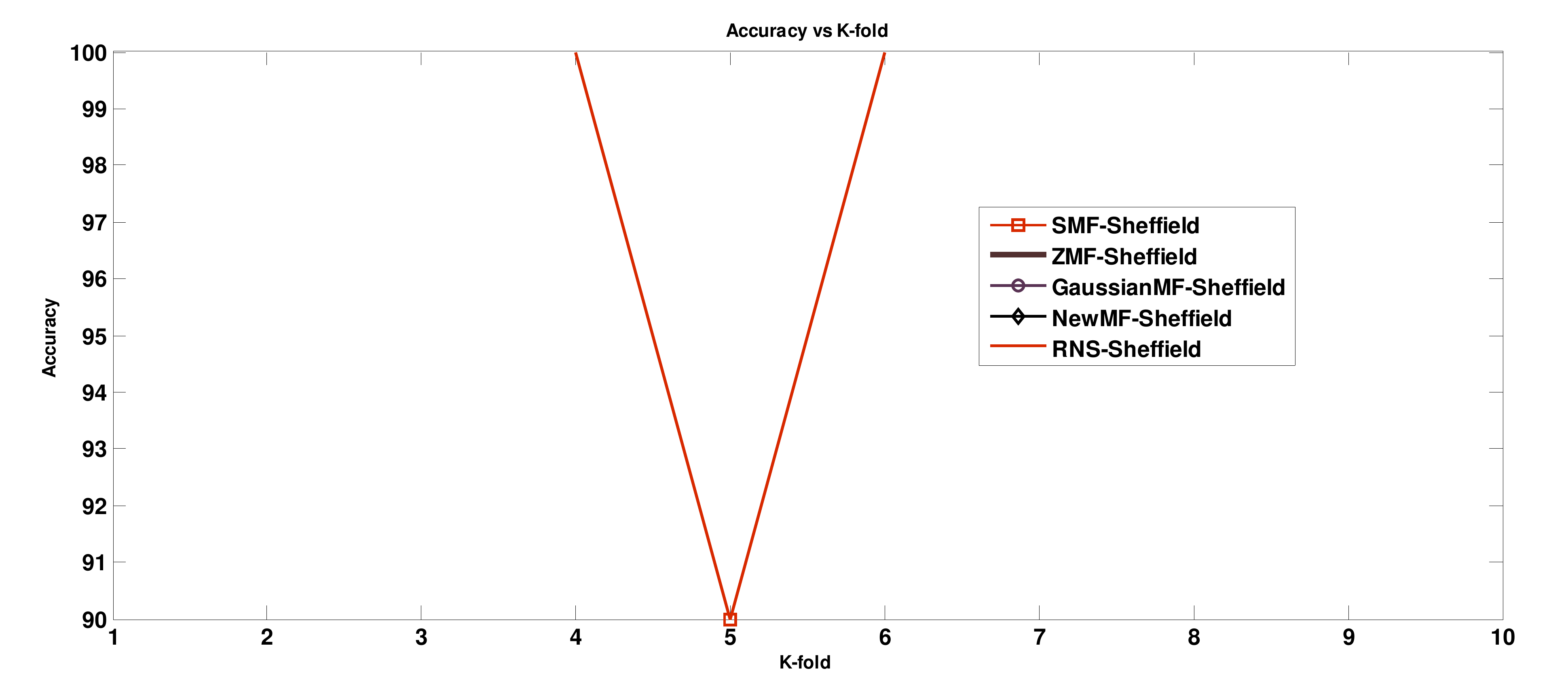}}
\caption{K-fold Recognition rates on Sheffield Database with 10 folds}
\label{fig:KFOLD_Sheffield}
\end{centering}
\end{figure*}

\begin{figure*}[t]
\begin{centering}
\centerline{\includegraphics[scale=0.3]{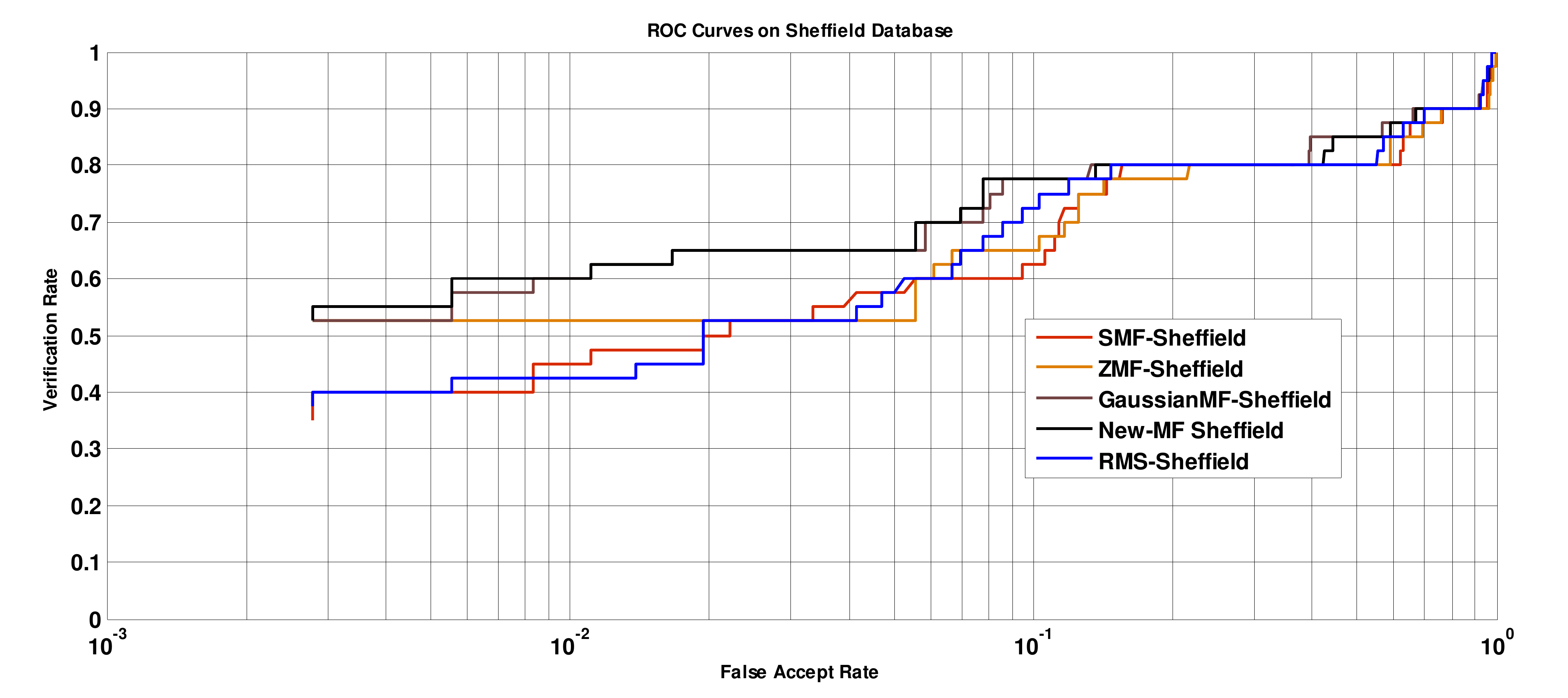}}
\caption{ROC Curves for various features on Sheffield database}
\label{fig:ROC_Sheffield}
\end{centering}
\end{figure*}

The experiment is also conducted on Sheffield Database. Firstly the images are resized to $90 \times 90$. After extracting the features for the complete database, the database is divided into train set and the test set. The first 50\% dataset is used for training and remaining 50\% is used for testing. For Sheffield database the recognition rate are given in Table \ref{tab:Sheffield}. The good performance is achieved from RMS features for the Degree of Polynomial 1,2, and the  second best performance is given by the New MF. With RMS features the recognition rate is 97.14\% with SVM poly 1.the weak performer is the S-MF as the SVM poly 2 has given a recognition rate of 85.71\%. The K-fold validation is done on the Sheffield Data base. Fig. \ref{fig:KFOLD_Sheffield} Shows the K-fold results of the recognition on Sheffield Database. The tabular representation is shown in the table 2 for the k-fold results. Database the best performances are  given by Z-MF, Gaussian-MF and New MF with the average of all ten folds as 100\% where as the  Root Mean Square (RMS) and Features obtained from S-MF with minimum recognition rate as 90\% and maximum recognition rate as 100\% and  average recognition rate of 99\%. Here in this case the New-MF has worked well. Whereas the RMS features are second best.   Receiver operating characteristic (ROC) curve is also plotted. For obtaining the ROC curves we have taken the K-NN Classifier. Fig. \ref{fig:ROC_Sheffield} Shows the ROC results of the recognition on Sheffield Database.
The recognition rates at 0.1 of FAR are as follows with S-MF the recognition rate we got is 62\%, with Z-MF the recognition rate 65\%, with Gaussian-MF and the New-MF recognition rate 78\%, and with RMS features the recognition rate obtained is 73\%. The better performance is given by Gaussian MF and the least performance is given by S-MF.

%
%

Comparisons with other works:
Arindam Kar et. al [12]
 $ \, $have reported the recognition rate on ORL database with Principle Component Analysis as 80.5\% and with Independent Component Analysis (ICA) with 85\%. Arindam Kar et. al in another work [11]
  have reported the recognition rate with Principle Component Analysis as 82.86\%. Author in [10]
   has reported recognition rate of 63\%  on Locally preserving projection (LPP). All the recognition are done with ORL database.

\section {CONCLUSIONS}
In this research work, we have developed five different Fuzzy Local Binary Patterns (Fuzzy LBP) as a local descriptor for face recognition. The features are designed keeping the aim to improve the recognition rate. Our main contributions include: The use of information set to LBP and to use it for face recognition to produce highly describable features. The information that is lost in computation of Basic LBP is now used for computation of final feature. In order to generate the optimal number of features, only non overlapping windows are considered, and hence computations of histogram are avoided. The proposed approach is found to be effective on images having variation in expressions, illumination and pose. The experiments reveal that better the representation of the uncertainty, better will be the recognition rates. Further work can be extended to use the type 2 fuzzy set, and design of new classifier.



{\bf {Abdullah Gubbi}} is currently working as an
Associate Professor in PA College of Engineering, Mangalore. He obtained his Bachelor of Engineering from Gulbarga University, Gulbarga.
He received his Masters degree in Electronics from
Walchand College of Engineering Sangli Maharashtra. His areas of interest include Image Processing, Pattern Recognition and VLSI Design. \\

{\bf{Mohammad Fazle Azeem}} is currently working as a Professor in AMU Aligarh (U.P). He obtained his Bachelor of Engineering from M.M.M. Engineering College, University Of Gorakhpur, Gorakhpur (U.P.) India. He received his Masters degree in Electrical Engineering from Aligarh Muslim University, Aligarh, India. He obtained his Ph.D., from Indian Institute of Technology, Delhi, New Delhi, India. His areas of interest include Control system, Image Processing and VLSI Design.\\

{\bf{Zahid Ahmed Ansari}}  is working as Professor in the department of Computer Science Engineering, P.A. College of Engineering, Mangalore, India. He received his M.E. degree from Birla Institute of Technology, Pilani, India and his Ph.D. degree from Jawaharlal Nehru Technological University, Kakinada, India. He has thirty research papers to his credit, published in various international journals and conferences. His areas of research include data mining, soft computing, high performance computing and model driven software development. He is a life member of CSI and also a member of ACM

\end{document}